\documentclass[a4paper,twoside]{article}

\usepackage{epsfig}
\usepackage{subcaption}
\usepackage{calc}
\usepackage{amssymb}
\usepackage{amstext}
\usepackage{amsmath}
\usepackage{amsthm}
\usepackage{multicol}
\usepackage{pslatex}
\usepackage{apalike}
\usepackage{algorithm2e}
\let\bibhang\relax
\usepackage{natbib}
\usepackage{multirow}
\usepackage{adjustbox}
\usepackage{xcolor}

\usepackage[bottom]{footmisc}
\usepackage{SCITEPRESS}     

\begin{document}

\title{From Observations to Causations: A GNN-based Probabilistic Prediction Framework for Causal Discovery}

\author{\authorname{Rezaur Rashid\sup{1}, Gabriel Terejanu\sup{2}}
\affiliation{Department of Computer Science, UNC Charlotte, Charlotte, NC, United States}
\email{{\sup{1}mrashid1@charlotte.edu}\\ {\sup{2}gabriel.terejanu@charlotte.edu}}
}



\abstract{Causal discovery from observational data is challenging, especially with large datasets and complex relationships. Traditional methods often struggle with scalability and capturing global structural information. To overcome these limitations, we introduce a novel graph neural network (GNN)-based probabilistic framework that learns a probability distribution over the entire space of causal graphs, unlike methods that output a single deterministic graph. Our framework leverages a GNN that encodes both node and edge attributes into a unified graph representation, enabling the model to learn complex causal structures directly from data. The GNN model is trained on a diverse set of synthetic datasets augmented with statistical and information-theoretic measures, such as mutual information and conditional entropy, capturing both local and global data properties. We frame causal discovery as a supervised learning problem, directly predicting the entire graph structure. Our approach demonstrates superior performance, outperforming both traditional and recent non-GNN-based methods, as well as a GNN-based approach, in terms of accuracy and scalability on synthetic and real-world datasets without further training. This probabilistic framework significantly improves causal structure learning, with broad implications for decision-making and scientific discovery across various fields.}

\onecolumn \maketitle \normalsize \setcounter{footnote}{0} \vfill

\section{\uppercase{Introduction}}
\label{introduction}

Causal inference from observational data is a fundamental task in many disciplines~\citep{koller2009probabilistic, pearl2019seven, peters2017elements, sachs2005causal, ott2003finding} and forms the backbone of many practical decision-making procedures as well as theoretical developments. Classical causal discovery algorithms test hypotheses of conditional independences to learn causal structure~\citep{spirtes2001causation}. Score-based causal discovery algorithms optimize fit scores over various graph structures~\citep{chickering2002optimal}. While effective in many situations, these approaches suffer from exponential run-times and combinatorial explosions in statistic complexity as the data sets grow~\citep{heckerman1995learning}. Advancements in machine learning, such as the NOTEARS algorithm, employ continuous optimization to enforce acyclicity, enhancing computational efficiency~\citep{zheng2018dags}. These approaches typically identify a single best causal graph rather than a probability distribution over multiple possible graphs, which can limit its ability to account for uncertainty in the causal discovery process.

The emergence of graph neural networks (GNNs) has revolutionized the field of predictive learning on graph-structured data, enabling powerful representations and insights from complex networks and relationships. From social network analysis to molecular property prediction~\citep{kipf2016semi, velickovic2017graph}, Graph Convolutional Networks (GCN) and other sophisticated variants such as Graph Attention Networks (GAT), have successfully exploited node and edge features to learn deep and hierarchical representations~\citep{zhou2020graph, waikhom2023survey}. Despite their success in areas such as network analysis and bioinformatics~\citep{hamilton2017inductive, lacerda2012discovering}, these methods have yet to be fully integrated into causal discovery frameworks. Such developments strongly motivate and justify the idea of utilizing GNNs for causal learning tasks~\citep{brouillard2020differentiable, peters2017elements}. For example, DAG-GNN~\citep{yu2019dag}, focuses on deterministic structure learning, while our methods use a probabilistic framework to better capture the inherent uncertainties in causal relationships. Furthermore,~\citet{li2020supervised} framed causal discovery as a supervised learning problem, directly predicting the entire DAG structure from observational data using neural networks. Similarly, the CausalPairs approach~\citep{fonollosa2019conditional, rashid2022causal} introduced a predictive framework for pairwise causal discovery.

Building on these advancements, this paper proposes a novel GNN-based probabilistic framework for causal discovery based on supervised learning that addresses the limitations of existing methods, including the work by~\citet{rashid2022causal} on causal pairs, by capturing global information directly from the data in the graph structure. 

Our work makes several key contributions: 
\begin{itemize}
    \item We introduce a novel probabilistic causal discovery framework based on GNNs that learns a probability distribution over causal graphs instead of producing a single deterministic graph.
    \item Our model is trained once on diverse synthetic datasets and can generalize to new datasets without requiring retraining, ensuring efficiency and broad applicability.
    \item We show that our approach performs better compared to traditional and recent causal discovery methods on both synthetic and real-world datasets.
\end{itemize}

Our approach surpasses benchmark methods, including traditional techniques: PC~\citep{spirtes2001causation}, GES~\citep{chickering2002optimal}; recent non-GNN-based methods: LiNGAM~\citep{shimizu2006linear}, NOTEARS-MLP~\citep{zheng2018dags}, DiBS~\citep{lorch2021dibs}, DAGMA~\citep{bello2022dagma}; and GNN-based method: DAG-GNN~\citep{yu2019dag}, in terms of accuracy on synthetic datasets generated from nonlinear structural equation models (SEMs), while also performing favorably compared to DAG-GNN and NOTEARS-MLP, and outperforming LiNGAM and GES for real-world dataset. 


The next section reviews the related work, followed by the problem formulation and a detailed explanation of our causal discovery approach using GNNs in the 'Methodology' section. The 'Experiments' section presents the empirical evaluation of our methods. Finally, the 'Conclusions' section summarizes our findings and discusses potential future improvements.

\section{\uppercase{Related Work}}
\label{relatedworks}

Structure learning from observational data typically follows either constraint-based or score-based methodologies. Constraint-based approaches, like the PC algorithm~\citep{spirtes2001causation}, start by employing conditional independence tests to map out the underlying causal graph's skeleton. Alternatively, score-based strategies, such as those implemented by GES~\citep{chickering2002optimal}, involve assigning scores to potential causal graphs according to specific scoring functions~\citep{bouckaert1993probabilistic, heckerman1995learning}, and then systematically exploring the graph space to identify the structure that optimizes the score~\citep{tsamardinos2006max, gamez2011learning}. However, the challenge of pinpointing the optimal causal graph is NP-hard, largely due to the combinatorial nature of ensuring acyclicity in the graph~\citep{mohammadi2015bayesian, mohan2012structured}. As a result, the practical reliability of these methods remains uncertain, especially when dealing with the complexities of real-world data.

Another approach focuses on identifying cause-effect pairs using statistical techniques from observational data. Fonollosa's work on the JARFO model~\citep{fonollosa2019conditional} is a notable effort in this direction to infer causal relationships from pairs of variables. Despite the promise of these pairwise methods, they often fail to leverage global structural information, limiting their effectiveness in constructing comprehensive causal graphs.

Recent advancements, such as the NOTEARS algorithm~\citep{notears}, incorporate continuous optimization techniques to ensure the acyclicity of the learned graph without requiring combinatorial constraint checks, representing a significant improvement in computational efficiency and scalability. However, experiments indicate that this method is highly sensitive to data scaling~\citep{reisach2021beware}. 

On the other hand, geometric deep learning, specifically GNNs, has revolutionized learning paradigms in domains dealing with graph-structured data~\citep{kipf2016semi, hamilton2017inductive, velickovic2017graph}. Despite the success of GNNs in various domains, their application in causal discovery is still emerging, but recent studies highlight rapid progress in both methodology and real-world impact~\citep{behnam2024graph, zhao2024causal, job2025exploring}. A few pioneering works have begun exploring this avenue, each with its own perspective~\citep{gao2024rethinking, zevcevic2021relating, singh2017deep}.~\citet{li2020supervised} propose a probabilistic approach for whole DAG learning using permutation equivariant models. This method demonstrates how supervised learning can be applied to structure discovery in graphs. ~\citet{lorch2022amortized} uses domain-specific supervised learning to generate inductive biases for causal discovery by characterizing all direct causal effects in that domain. DAG-GNN~\citep{yu2019dag} uses a variational autoencoder parameterized by GNNs to learn directed acyclic graphs (DAGs), focusing on deterministic structure learning and primarily utilizing node features. Our methods, in contrast, emphasize a probabilistic framework, incorporating both node and edge features. Interestingly, our algorithm can complement DAG-GNN by providing a probabilistic distribution over possible DAGs, potentially refining its causal structure learning. Another study presents a gradient-based method for causal structure learning with a graph autoencoder framework, accommodating nonlinear structural equation models and vector-valued variables, and outperforming existing methods on synthetic datasets~\citep{ng2019graph}. Furthermore, the Gem framework provides model-agnostic, interpretable explanations for GNNs by formulating the explanation task as a causal learning problem, achieving superior explanation accuracy and computational efficiency compared to state-of-the-art alternatives~\citep{lin2021generative}. 

Despite promising advances, existing methods have yet to fully exploit the capabilities of GNNs for causal discovery, particularly in modeling complex causal structures from observational data in a scalable and uncertainty-aware manner. Many prior approaches either focus on deterministic outputs or omit edge-level features and probabilistic modeling, limiting their ability to generalize. Compared to traditional algorithms like PC, which iteratively apply conditional independence tests to construct a causal graph for each dataset, our framework predicts a probability distribution over DAGs directly from feature-rich edge representations using a GNN. This predictive shift enables generalization across datasets, removes the need for dataset-specific optimization, and allows for uncertainty quantification. Unlike DAG-GNN and NOTEARS, which optimize a structure per instance, our method is trained once and can infer causal graphs in a single forward pass. As noted by~\citet{jiang2023graph}, GNN-based causal discovery remains underexplored, especially in probabilistic settings, a gap our work seeks to fill.

\section{\uppercase{Methodology}}
\label{methodology}

Assuming we have $n$ i.i.d. observations in the data matrix $\mathbf{X} = [\mathbf{x_1}\ldots\mathbf{x_d}] \in \mathbb{R}^{n\times d}$, causal discovery attempts to estimate the underlying causal relations encoded by the di-graph, $\mathcal{G}=(V,E)$. $V$ consists of nodes associated with the observed random variables $X_i$ for $i=1 \ldots d$ and the edges in $E$ represent the causal relations encoded by $\mathcal{G}$. In other words, the presence of the edge $i \to j$ corresponds to a direct causal relation between $X_i$ (cause) and $X_j$ (effect).

Our approach uses a graph neural network model to predict the probability $p(e_{ij}|f)$ of an edge $e_{ij}$ between nodes $X_i$ and $X_j$ given their feature representations.
\begin{eqnarray}
    \label{eq:model}
    p(e_{ij} | \mathbf{h}_i, \mathbf{h}_j, \mathbf{e}_{ij}) &=& f([\mathbf{h}_i, \mathbf{h}_j, \mathbf{e}_{ij}]),~\text{for}~i<j
\end{eqnarray}

\noindent
Here, 
\begin{itemize}
    \item $\mathbf{h}_i$ and $\mathbf{h}_j$ represent the feature vectors of nodes $X_i$ and $X_j$ after the GNN's message passing and aggregation operations.
    \item $\mathbf{e}_{ij}$ represents the feature vector of the edge $e_{ij}$  between nodes $X_i$ and $X_j$.
    \item $[\mathbf{h}_i, \mathbf{h}_j, \mathbf{e}_{ij}]$ denotes the concatenation of the feature vectors of nodes $X_i$ and $X_j$ and the edge features $\mathbf{e}_{ij}$.
    \item The function $f$ represents the GNN classifier that outputs the probability $p(e_{ij} | \mathbf{h}_i, \mathbf{h}_j, \mathbf{e}_{ij})$ of there being an edge $e_{ij} \in [-1, 0, 1]$.
\end{itemize}

\begin{equation}
\nonumber
\small
\label{eq-causal label}
e_{ij} = 
    \begin{cases}
    -1: &\text{$j \to i$, causal relation exists from $X_j$ to $X_i$}\\
    \phantom{-}0: &\text{$i \not\to j$ and $j \not\to i$,} \\
    & \text{no direct causal relation between $X_i$ and $X_j$}\\
    \phantom{-}1: &\text{$i \to j$, causal relation exists from $X_i$ to $X_j$}
    \end{cases}
\end{equation}

\begin{figure*}[h!]
  \centering
  \includegraphics[width=0.85\linewidth]{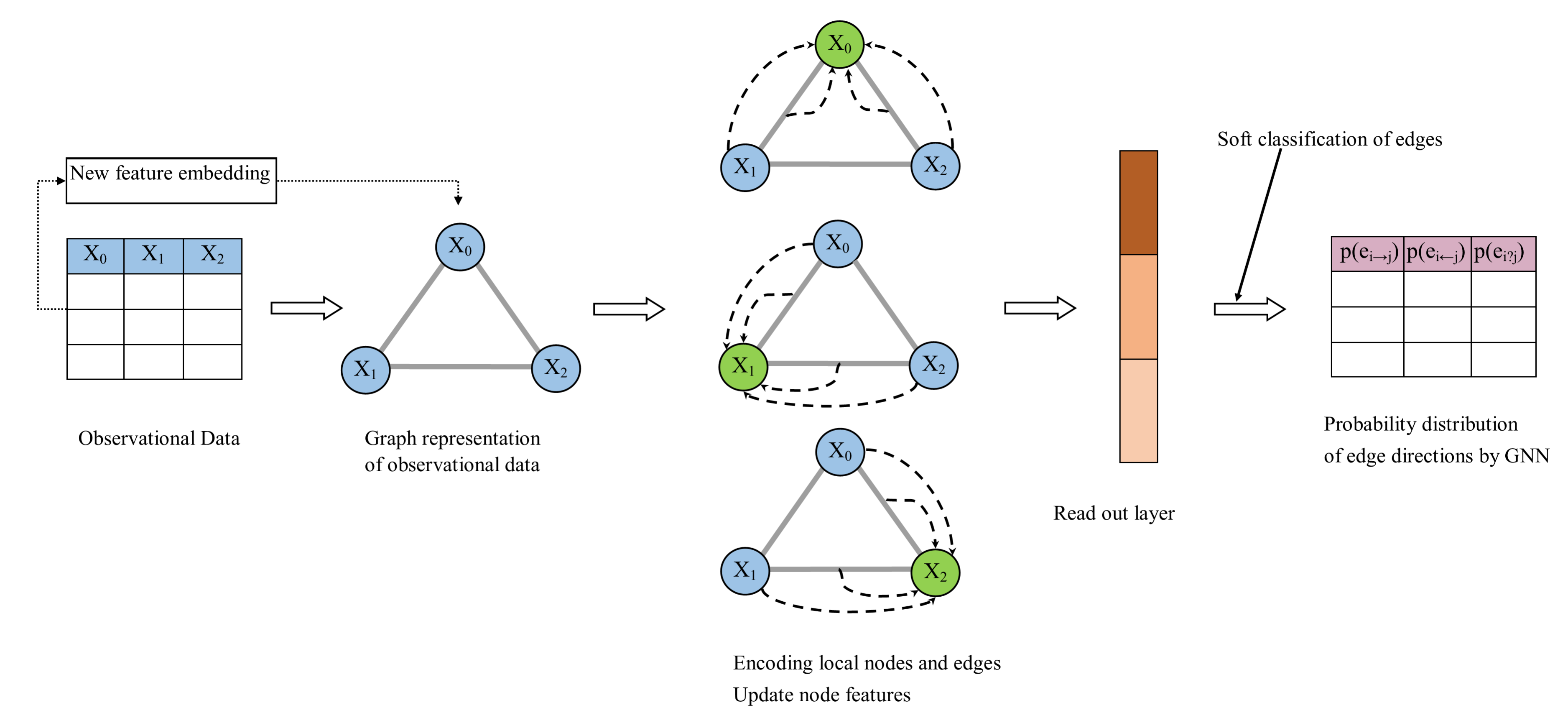}

  \caption{Schematic of the proposed framework. Each node is initialized with statistical features, and each edge with aggregated information-theoretic, statistical, and causal-pairs features~\citep{rashid2022causal}. The GNN predicts edge directions, capturing both local and global dependencies to infer the underlying causal graph.}
  \label{fig:gnn-model}
\end{figure*}

\subsection{Feature Engineering and Graph Construction}

We first construct a fully connected graph $\mathcal{G}=(V,E)$, where $V$ is the set of all attributes in the observational dataset, and $E$ is the set of edges between nodes (attributes) such that every node is connected with every other node which leads to $d(d-1)/2$ edges in the graph for a dataset with $d$ attributes. We then extract statistical and information-theoretic measures on the attributes in the observational dataset to represent each node with 13 features and each edge with 114 features between node pairs in the graph. 

Node features encode statistical properties such as entropy, skewness, and kurtosis, summarizing the distribution of each variable. Edge features aggregate information-theoretic and statistical relationships between variable pairs (e.g., mutual information, conditional entropy, polynomial fit error, Pearson correlation) to capture both linear and nonlinear dependencies. We also incorporate the probability distribution over the edge direction using the causal-pairs model~\citep{rashid2022causal} as 3 additional edge features, resulting in a total of 114 edge features per edge in the graph. A complete list of all node and edge features can be found in Appendix~\ref{appendix:features}. 

A simplified illustration is shown in Figure~\ref{fig:gnn-model}. The intuition behind this approach is that by creating a comprehensive feature set that includes both node and edge features, we can capture a rich representation of the underlying dependencies and interactions between variables. The fully connected graph ensures that all possible relationships are considered, allowing the model to learn from a wide range of potential causal connections. Furthermore, incorporating the probability distribution from the causal-pairs model adds another layer of probabilistic reasoning, enhancing the model's ability to infer causal directions accurately. This multi-faceted feature representation enables the GNN to leverage both local and global information, leading to more accurate and reliable causal predictions.

\subsection{Developing the Graph Neural Network (GNN) Model}

Graph neural networks (GNNs) are a family of architectures that leverage graph structure, node features, and edge features to learn dense graph representations. GNNs employ a neighborhood aggregation strategy, iteratively updating node representations by aggregating information from neighboring nodes. For example, a basic operator for neighborhood information aggregation is the element-wise mean.

In our study, we utilize a GNN model as an edge classifier by training it on synthetic datasets with underlying causal graphs to infer the probability distribution over edge directions through supervised learning. Although recent works propose more sophisticated GNN variants, we specifically adopt GraphSAGE as our backbone due to its scalability and efficient sampling-based message passing, which is particularly well-suited for large, fully connected graphs. This choice strikes a balance between computational efficiency, ease of implementation, and empirical robustness, rather than architectural novelty.

Starting with a fully connected complete graph, GraphSAGE enables efficient learning by sampling and aggregating messages from a subset of neighbors, improving scalability in message-passing iterations without compromising model accuracy. This aligns with our intuition regarding the importance of local neighborhoods in characterizing conditional independences - a key aspect of causal discovery. Although GraphSAGE is primarily designed to update node features based on neighboring node features, we extend it to incorporate edge features into the message-passing process, allowing the model to better capture pairwise dependencies relevant to causal inference. The model learns a mapping from the edge features (e.g., mutual information, conditional entropy) to edge direction probabilities, using training graphs with known causal structure. This replaces the need for dataset-specific search or constraint satisfaction.

To integrate both node and edge features, we define the message \( m_{uv}^{(k)} \) as a combination of the feature vectors of nodes \( u \) and \( v \) at layer (k-1), along with the edge feature vector \( e_{uv} \). The updated equations for message passing and node feature updates are as follows:

\begin{equation}
\label{message_calculation}
    m_{uv}^{(k)} = \text{CONCAT}(h_u^{(k-1)}, h_v^{(k-1)}, e_{uv})
\end{equation}

\begin{equation}
\label{message_aggregation}
m_v^{(k+1)} = \frac{1}{|N(v)|} \sum_{u \in N(v)} m_{uv}^{(k)}
\end{equation}

\begin{equation}
\label{node_update}
h_v^{(k+1)} = \sigma \left( W \cdot \text{CONCAT}(h_v^{(k)}, m_v^{(k+1)}) \right)
\end{equation}

Here,
\begin{itemize}
    \item For each neighboring node \( u \) of node \( v \), we calculate a message \( m_{uv}^{(k)} \) by concatenating the feature vectors of node \( u \) and node \( v \) at layer \( k-1 \) along with the edge feature vector \( e_{uv} \).
    \item The messages \( m_{uv}^{(k)} \) from all neighbors \( u \in N(v) \) are aggregated by summing them and normalizing by the number of neighbors \( |N(v)| \). This normalization ensures that contributions from all neighbors are equally weighted.
    \item The aggregated message \( m_v^{(k+1)} \) is concatenated with the current feature vector of node \( v \) (\( h_v^{(k)} \)).
    \item The concatenated vector is then passed through a linear transformation defined by the learnable weight matrix \( W \), followed by a non-linear activation function \( \sigma \) (e.g., ReLU).
\end{itemize}

This model captures both local and global dependencies in the graph structure, enhancing the accuracy of inferred causal relations between nodes considering their relationships with neighbors. After multiple rounds of message passing, the final node embeddings represent each node and edge in the graph, allowing for the prediction of edge direction probabilities (forward, reverse, or no edge) between any pair of nodes.

\subsection{Probabilistic Inference}

The edge probabilities predicted by the GNN model define a distribution over all possible graphs, rather than directly yielding a single acyclic structure  $p(\mathcal{G}_{DAG})$. This probabilistic formulation captures the inherent uncertainty in causal relationships, allowing for a more comprehensive representation of potential causal structures instead of committing to a single deterministic graph.

To extract meaningful graph representations from this probabilistic space, we consider four approaches as presented in~\citet{rashid2022causal}: (1) Probability of Graph (PG), which represents the full probability distribution over directed graphs and can be used to sample a digraph; (2) Maximum Likelihood Digraph (MLG), which selects the most probable edge directions to form a representative structure; (3) Probability of DAG (PDAG), which refines the probability distribution by incorporating acyclicity constraints and enables sampling of DAGs; and (4) Maximum Likelihood DAG (MLDAG), which provides a deterministic estimate of the most probable acyclic structure. The transition from PG/MLG to PDAG/MLDAG is crucial: while the first two approaches allow cycles, the latter two explicitly enforce the acyclicity assumption required for valid causal graphs. These methods progressively refine the estimated causal graph, ensuring structural validity while balancing probabilistic inference with computational efficiency. This probabilistic formulation supports multiple inference strategies, enabling both flexible sampling and strict acyclicity enforcement. It contrasts with deterministic methods like PC or GES, which return only a single output graph without uncertainty estimates and require full recomputation per dataset. For clarity, we briefly outline each approach below and refer to~\citet{rashid2022causal} for detailed algorithmic derivations and proofs.

\paragraph{Sample Digraph (PG).} 
The first and most intuitive approach is to construct the probability distribution of a digraph $\mathcal{G}$ using the maximum entropy principle. After computing the probability distributions of causal relationships between node pairs or edge directions, this method assumes that edge directions are independent, resulting in a straightforward formulation (Eq.~\ref{eq:PG}).
\begin{equation}
\label{eq:PG}
    p(\mathcal{G}|f) = \prod_{i<j} p(e_{ij}|f) 
\end{equation}

\paragraph{Maximum Likelihood Digraph (MLG).} Given the above naive distribution over digraphs, one can extract a single representative structure by selecting the edge directions with the highest probabilities. This leads to the maximum likelihood digraph, which represents the most likely structure according to Eq.~\ref{eq:MLG}.
\begin{equation}
\label{eq:MLG}
    \mathcal{G}_\text{ML} = \arg\max_\mathcal{G} p(\mathcal{G}|f)  
\end{equation}

Note that the samples from the probability distribution, Eq.~\ref{eq:PG}, and the  maximum likelihood digraph in Eq.~\ref{eq:MLG}, are digraphs with no guarantees of acyclicity. 

\paragraph{Sample DAG (PDAG).} A more principled approach refines the naive distribution by explicitly ensuring acyclicity of the generated graphs. Rather than independently sampling edge directions, this method incorporates DAG constraints by marginalizing over the topological ordering $\pi$ of vertices, as shown in Eq.~\ref{eq:PDAG}:
\begin{equation}
\label{eq:PDAG}
    p(\mathcal{G}|f, \text{DAG}) = \sum_\pi p(\mathcal{G}|f, \text{DAG}, \pi) p(\pi|f)
\end{equation}

Due to the computational intractability of marginalizing over $\pi$, we approximate the probability of DAGs by conditioning on the maximum likelihood topological ordering, $\pi_{\text{ML}}$. This leads to the following approximation:
\begin{equation}
\label{eq:PDAG_new}
    p(\mathcal{G}|f, \text{DAG}, \pi_\text{ML}) = \prod_{\pi^{-1}_\text{ML}[i] < \pi^{-1}_\text{ML}[j]} p(e_{i\rightarrow j}|f) 
\end{equation}

Furthermore, we approximate the maximum likelihood topological ordering, $\pi_{\text{ML}}$, by performing a topological sort on the Maximum Spanning DAG (MSDAG)~\citep{schluter-2014-maximum}, which is computed from the induced weighted graph $\mathcal{G}_A$, defined by the probabilities of causal relations:
\begin{eqnarray}
    \pi_\text{ML} = \arg\max_\pi p(\pi|f) 
    \approx \text{toposort(MSDAG($\mathcal{G}_A$))}
\end{eqnarray}

In practice, to compute the topological ordering from the MSDAG of $\mathcal{G}_A$, we follow the procedure introduced by \citet{mcdonald2006online}: first constructing a maximum spanning tree, then incrementally adding edges in descending order of weights while ensuring acyclicity at each step.

\paragraph{Maximum Likelihood DAG (MLDAG).} Extending the MLG approach to enforce acyclicity, the maximum likelihood DAG provides a deterministic representation of the most probable causal structure. Instead of selecting the highest-probability edges independently, this method ensures acyclicity by incorporating the DAG constraints introduced in the previous approach. In other words, edges are selected in order of probability, but only if they do not introduce a cycle with respect to the current partial ordering. Thus the resulting graph is constructed by selecting the most probable edges while maintaining a valid topological ordering, as formulated in Eq.~\ref{eq:MLDAG}.
\begin{eqnarray}
    \mathcal{G}_\text{DAG}     &\approx&  \arg\max_\mathcal{G} p(\mathcal{G}|f, \text{DAG}, \pi_\text{ML}) \label{eq:MLDAG}
\end{eqnarray}


\section{\uppercase{Experiments}}
\label{experiments}

To evaluate the effectiveness of our proposed probabilistic inference methods, we conduct experiments on synthetic, benchmark, and real-world datasets. We compare our approaches, GNN-PG (sample digraph from the probability distribution), GNN-MLG (maximum likelihood estimate digraph), GNN-PDAG (sample DAG from the probability distribution), and GNN-MLDAG (DAG using the maximum likelihood estimate), against both traditional and recent causal discovery methods.

Specifically, we benchmark our methods against classical algorithms such as PC~\citep{spirtes2001causation} and GES~\citep{chickering2002optimal}, as well as recent approaches including LiNGAM~\citep{shimizu2006linear}, DAG-GNN~\citep{yu2019dag}, NOTEARS-MLP~\citep{notears}, DiBS~\citep{lorch2021dibs}, and DAGMA~\citep{bello2022dagma}. For PC, GES, and LiNGAM, we use publicly available implementations\footnote{PC: https://github.com/keiichishima/pcalg\label{fn-pc}}\textsuperscript{,}\footnote{GES: https://github.com/juangamella/ges\label{fn-ges}}\textsuperscript{,}\footnote{LiNGAM: https://lingam.readthedocs.io/en/latest\label{fn-lingam}}, while for DAG-GNN, NOTEARS-MLP, DiBS, and DAGMA, we follow the implementations provided by the respective authors\footnote{DAG-GNN: https://github.com/fishmoon1234/DAG-GNN\label{fn-daggnn}}\textsuperscript{,}\footnote{NOTEARS-MLP: https://github.com/xunzheng/notears\label{fn-notears}}\textsuperscript{,}\footnote{DiBS: https://github.com/larslorch/dibs\label{fn-dibs}}\textsuperscript{,}\footnote{DAGMA: https://github.com/kevinsbello/dagma\label{fn-dagma}}. We use default hyperparameter settings for all methods to ensure a fair comparison.

\subsection{Datasets}

\paragraph{Synthetic Data} We generated synthetic data to train our GNN model on causal graph estimation, producing 200 graphs with 72 different combinations of nodes ($d=[10,20,50,100]$), edges ($e=[1d, 2d, 4d]$), data samples per node ($n=[500, 1000, 2000]$), and graph models (Erdos-Renyi and Scale-Free). Non-linear data samples were generated similarly to the NOTEARS-MLP implementation, with random graph structures and ground truth for training. The process for generating synthetic test data follows the methodology outlined in~\citet{rashid2022causal}, where 160 types of graph combinations were considered, each with varying numbers of nodes, edges, graph types, and data samples per node. 

\paragraph{CSuite Data} In addition to our synthetic test datasets, we employed five benchmark datasets from Microsoft CSuite, a collection designed for evaluating causal discovery and inference algorithms~\citep{geffner2022deep}. The CSuite data is generated from well-defined hand-crafted structural equation models (SEMs), which serve to test various aspects of causal inference methodologies. The five datasets utilized in our study are:
\textit{large\_backdoor} (9 nodes, 10 edges); \textit{weak\_arrows} (9 nodes, 15 edges); \textit{mixed\_simpson} (4 nodes, 4 edges); \textit{nonlin\_simpson} (4 nodes, 4 edges); \textit{symprod\_simpson} (4 nodes, 4 edges);. Each dataset includes 6000 data samples, and a corresponding ground truth graph, providing a basis for performance evaluation.

\paragraph{Real-World Data} We used the dataset from~\citet{sachs2005causal}, based on protein expression levels. This dataset is widely used due to its consensus ground truth of the graph structure, consisting of 11 protein nodes and 17 edges representing the protein signaling network. We aggregated 9 data files, resulting in a sample size of 7466 for our experiments.

\subsection{Metrics}

We evaluate the quality of the inferred causal graphs using True Positive Rate (TPR), False Positive Rate (FPR), and Structural Hamming Distance (SHD). A lower SHD and FPR indicate better performance, while a higher TPR is preferable. To ensure a fair comparison, these metrics are computed consistently across all methods, following the procedures used in prior evaluations of PC, GES, and NOTEARS-MLP. GNN-based and CausalPairs-based methods adhere to the implementation framework described in~\citet{rashid2022causal}.

\subsection{Results}
\label{results}

Table~\ref{table:sf-er} showcases the performance of our GNN-based methods on 80 Scale-Free (SF) and 80 Erdos-Renyi (ER) graph structures. Our methods consistently outperform traditional and recent approaches, demonstrating improved recovery of causal structures through reduced Structural Hamming Distance (SHD) and increased True Positive Rate (TPR).
Key observations across both graph structures include:

\begin{enumerate}
    \item Our GNN-based methods, especially GNN-PDAG and GNN-MLDAG, consistently achieve lower SHD and higher TPR values compared to CausalPairs methods; traditional methods such as PC and GES; and DiBS. They also perform favorably or better than advanced methods such as LiNGAM, DAG-GNN, NOTEARS-MLP, and DAGMA. Notably, they significantly improve TPR while maintaining low SHD.

    \item The GNN-MLG method significantly minimizes false positive causal relationships, but at the cost of a lower TPR. Other GNN-based methods balance TPR and FPR.

    \item Enforcing DAG constraints in GNN-PDAG and GNN-MLDAG improves performance metrics relative to GNN-PG and GNN-MLG, highlighting the benefit of integrating global structural information to enhance accuracy.

\end{enumerate}

\begin{figure*}
  \centering

  \begin{tabular}{@{}c@{}}
    \includegraphics[width=0.99\linewidth, trim = 0 0 85 0, clip]{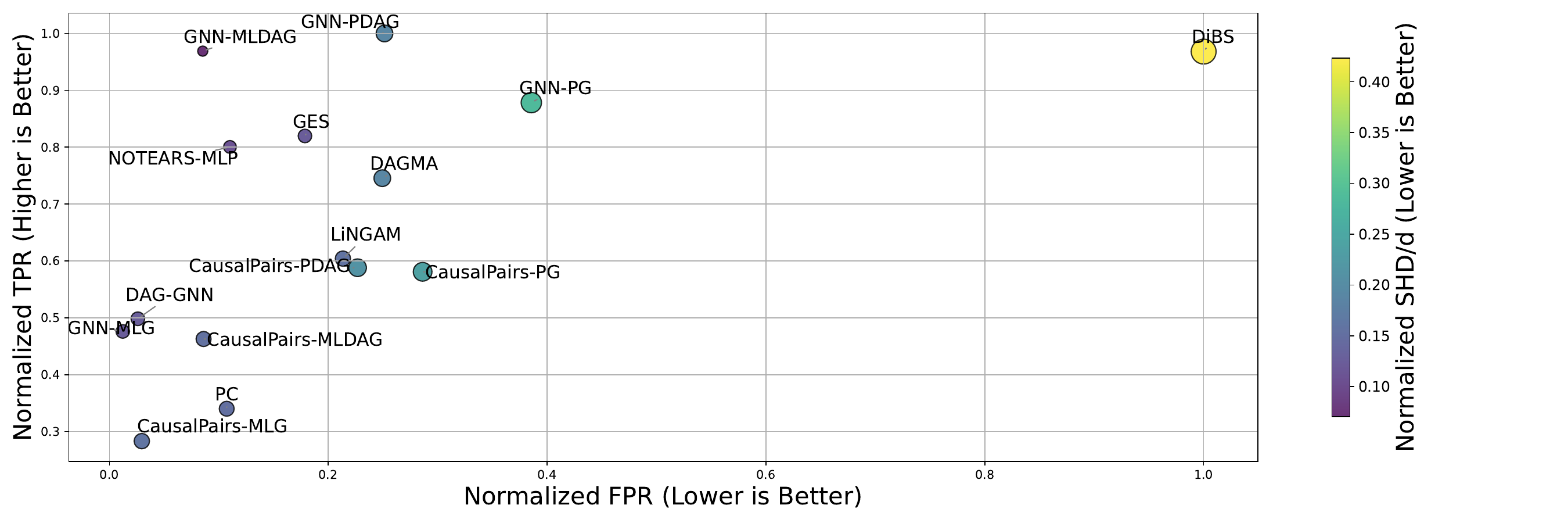} \\[\abovecaptionskip]
    \small (a) Node : Edge (1:1)
  \end{tabular}

  \vspace{5pt}

  \begin{tabular}{@{}c@{}}
    \includegraphics[width=0.99\linewidth, trim = 0 0 85 0, clip]{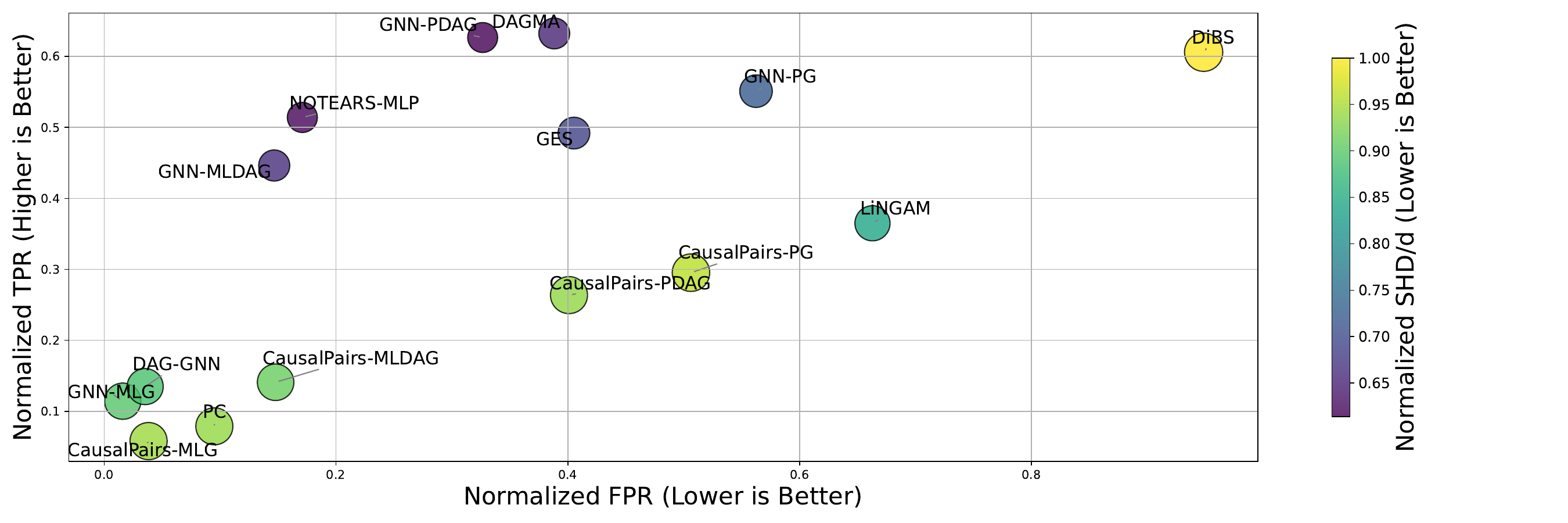} \\[\abovecaptionskip]
    \small (b) Node : Edge (1:4)
  \end{tabular}

    \caption{Comparison of normalized Structural Hamming Distance (SHD/d), True Positive Rate (TPR), and False Positive Rate (FPR) across methods on Erdos-Renyi (ER) and Scale-Free (SF) graphs, evaluated for both sparse (1:1) and dense (1:4) node-to-edge ratios. Metrics are computed as the mean and standard error over 80 randomly generated graphs for each condition.}
    \label{fig:results}
\end{figure*}

Figure~\ref{fig:results} presents a comprehensive comparison of the Structural Hamming Distance (SHD), True Positive Rate (TPR), and False Positive Rate (FPR) performance metrics for different methods on 160 SF and ER graphs with node-to-edge ratios of 1:1 and 1:4. 

Our GNN-based methods, specifically GNN-PDAG and GNN-MLDAG, consistently achieve lower SHD values than traditional methods (PC and GES), CausalPairs methods, and advanced methods (NOTEARS-MLP, DAG-GNN, and DAGMA). Notably, our proposed methods (GNN-PG, GNN-PDAG, and GNN-MLDAG) demonstrate significantly higher TPRs than all other methods, indicating improved accuracy in identifying true causal relationships. GNN-PDAG and GNN-MLDAG exhibit robust performance across both sparse (1:1) and dense (1:4) graphs, showcasing their ability to accurately recover causal structures with fewer errors. The improvement is more pronounced in denser graphs (1:4 node-to-edge ratio), showing promise in handling complex, highly connected networks. 


\begin{table*}[htpb]
\centering
\fontsize{9}{11}\selectfont
\caption{Comparison of edge probability model trained on GNN framework. The means and standard errors of the performance metrics are based on the 80 Scale-Free (SF) and 80 Erdos-Renyi (ER) graph structures in the test data.}
\begin{adjustbox}{width=0.80\linewidth}
\begin{tabular}{|l|r|r|r|r|r|r|}
\hline
{Dataset Name $\rightarrow$} & \multicolumn{3}{c|}{{Scale-Free (SF)}} & \multicolumn{3}{c|}{{Erdos-Renyi (ER)}} \\
\cline{1-7}
{Method $\downarrow$ | Metrics $\rightarrow$ }& {SHD/d} & {TPR} & {FPR} & {SHD/d} & {TPR} & {FPR} \\
\hline
GNN PG & 1.88$\pm$0.08 & 0.51$\pm$0.02 & 0.30$\pm$0.01 & 2.08$\pm$0.11 & 0.52$\pm$0.02 & 0.52$\pm$0.06 \\ \hline
GNN MLG & 1.85$\pm$0.13 & 0.20$\pm$0.02 & 0.01$\pm$0.00 & 2.17$\pm$0.17 & 0.25$\pm$0.02 & 0.01$\pm$0.00 \\ \hline
GNN PDAG & 1.55$\pm$0.07 & 0.56$\pm$0.02 & 0.19$\pm$0.01 & 1.75$\pm$0.11 & 0.61$\pm$0.03 & 0.28$\pm$0.03 \\ \hline
GNN MLDAG & 1.40$\pm$0.11 & 0.48$\pm$0.03 & 0.08$\pm$0.01 & 1.66$\pm$0.15 & 0.54$\pm$0.03 & 0.13$\pm$0.02 \\ \hline
CausalPairs PG & 2.02$\pm$0.12 & 0.31$\pm$0.01 & 0.26$\pm$0.02 & 2.38$\pm$0.14 & 0.39$\pm$0.02 & 0.72$\pm$0.10 \\ \hline
CausalPairs MLG & 1.97$\pm$0.13 & 0.12$\pm$0.01 & 0.03$\pm$0.01 & 2.32$\pm$0.17 & 0.15$\pm$0.02 & 0.07$\pm$0.01 \\ \hline
CausalPairs PDAG & 1.96$\pm$0.12 & 0.30$\pm$0.01 & 0.21$\pm$0.02 & 2.30$\pm$0.15 & 0.38$\pm$0.02 & 0.61$\pm$0.09 \\ \hline
CausalPairs MLDAG & 1.88$\pm$0.13 & 0.20$\pm$0.01 & 0.09$\pm$0.01 & 2.18$\pm$0.16 & 0.28$\pm$0.02 & 0.29$\pm$0.05 \\ \hline
PC & 1.93$\pm$0.15 & 0.17$\pm$0.02 & 0.08$\pm$0.01 & 2.40$\pm$0.21 & 0.17$\pm$0.02 & 0.22$\pm$0.04 \\ \hline
GES & 1.43$\pm$0.11 & 0.51$\pm$0.03 & 0.26$\pm$0.04 & 1.78$\pm$0.13 & 0.48$\pm$0.02 & 0.87$\pm$0.15 \\ \hline
LiNGAM & 1.68$\pm$0.11 & 0.35$\pm$0.02 & 0.34$\pm$0.04 & 1.97$\pm$0.13 & 0.43$\pm$0.02 & 1.04$\pm$0.17 \\ \hline
DAG-GNN & 1.75$\pm$0.12 & 0.24$\pm$0.02 & 0.02$\pm$0.00 & 2.10$\pm$0.17 & 0.27$\pm$0.02 & 0.06$\pm$0.00 \\ \hline
NOTEARS & 1.36$\pm$0.11 & 0.47$\pm$0.02 & 0.12$\pm$0.02 & 1.33$\pm$0.10 & 0.58$\pm$0.02 & 0.32$\pm$0.06 \\ \hline
DiBS & 2.51$\pm$0.08 & 0.32$\pm$0.02 & 0.91$\pm$0.25 & 2.78$\pm$0.10 & 0.34$\pm$0.02 & 0.38$\pm$0.06 \\ \hline
DAGMA & 1.39$\pm$0.09 & 0.54$\pm$0.02 & 0.21$\pm$0.02 & 1.80$\pm$0.11 & 0.51$\pm$0.02 & 0.65$\pm$0.10 \\
\hline
\end{tabular}
\end{adjustbox}
\label{table:sf-er}
\end{table*}


Tables~\ref{table:csuite} present the results of applying our methods to five datasets from the Microsoft CSuite. Our methods achieve significantly lower SHD, higher TPR, and lower FPR compared to all other methods, demonstrating the robustness and generalizability of our GNN-based framework across diverse datasets. Compared to the synthetic datasets presented in Table~\ref{table:sf-er}, the Microsoft CSuite datasets have fewer nodes and edges. Additionally, the three smaller datasets from Microsoft CSuite allow us to demonstrate the method's capability to recover various graph structures learned directly from data.

In these datasets, which include graphs with four nodes and four edges, our methods accurately identified $V$ structures such as $A \rightarrow B \leftarrow C$. This ability to capture fork or collider structures highlights the method's precision in determining causal directions and understanding interactions between variables. We also observed that in datasets like mixed\_simpson and nonlin\_simpson, with confounder structures such as $B \leftarrow A \rightarrow C$, our methods demonstrated the ability to recognize common causes affecting multiple outcomes. Chain structures like $A \rightarrow B \rightarrow C$ were also accurately recovered, showcasing the capability to model sequential causal relationships. For instance, among two of these datasets, our GNN-based methods achieved a SHD score of 0 and a TPR score of 1, perfectly identifying the true graph, and validating our methods' effectiveness in learning complex causal structures.


\begin{figure*}[htpb]
  \centering
  \begin{tabular}{@{}c@{}c@{}c@{}}  
    {\includegraphics[width=.32\linewidth]{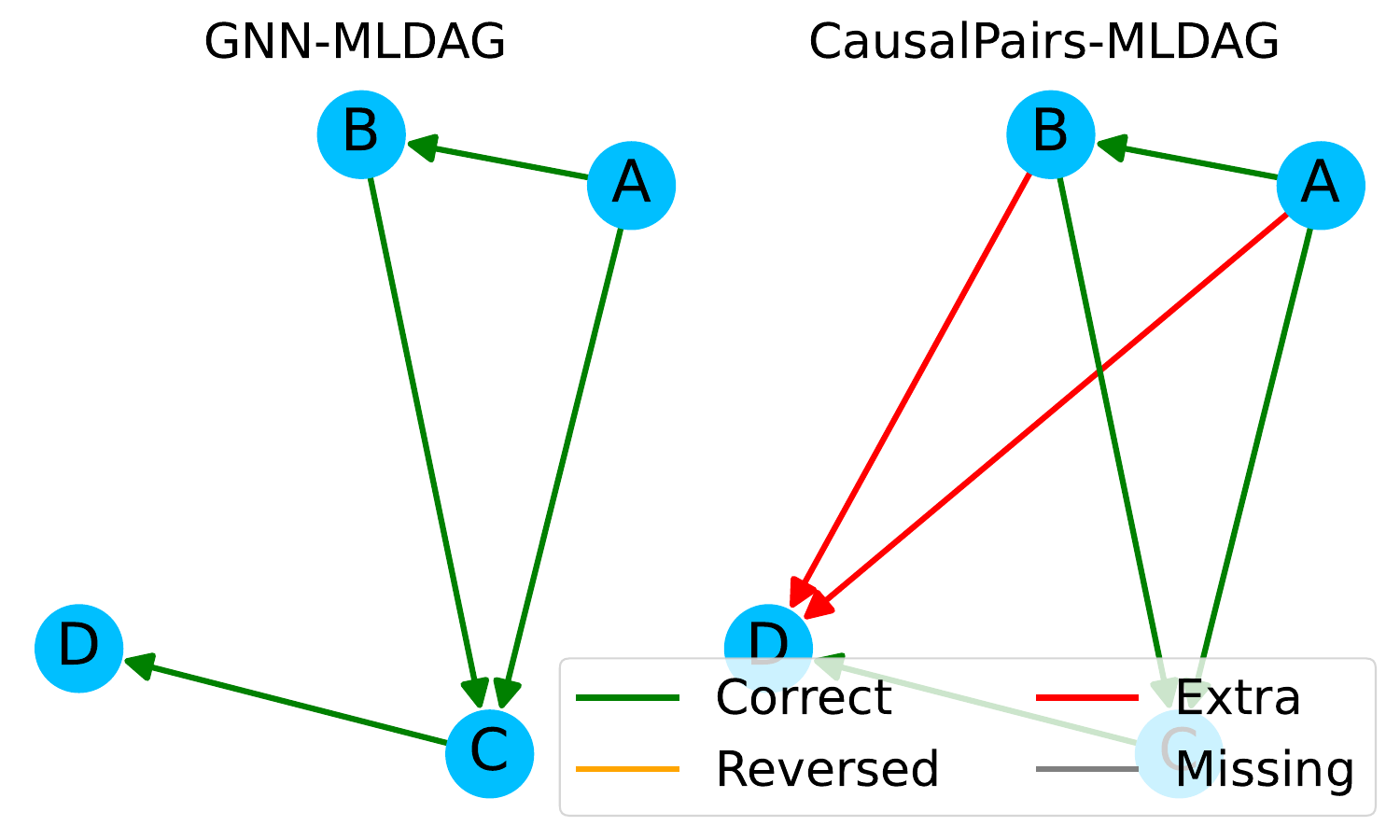}} &
    {\includegraphics[width=.32\linewidth]{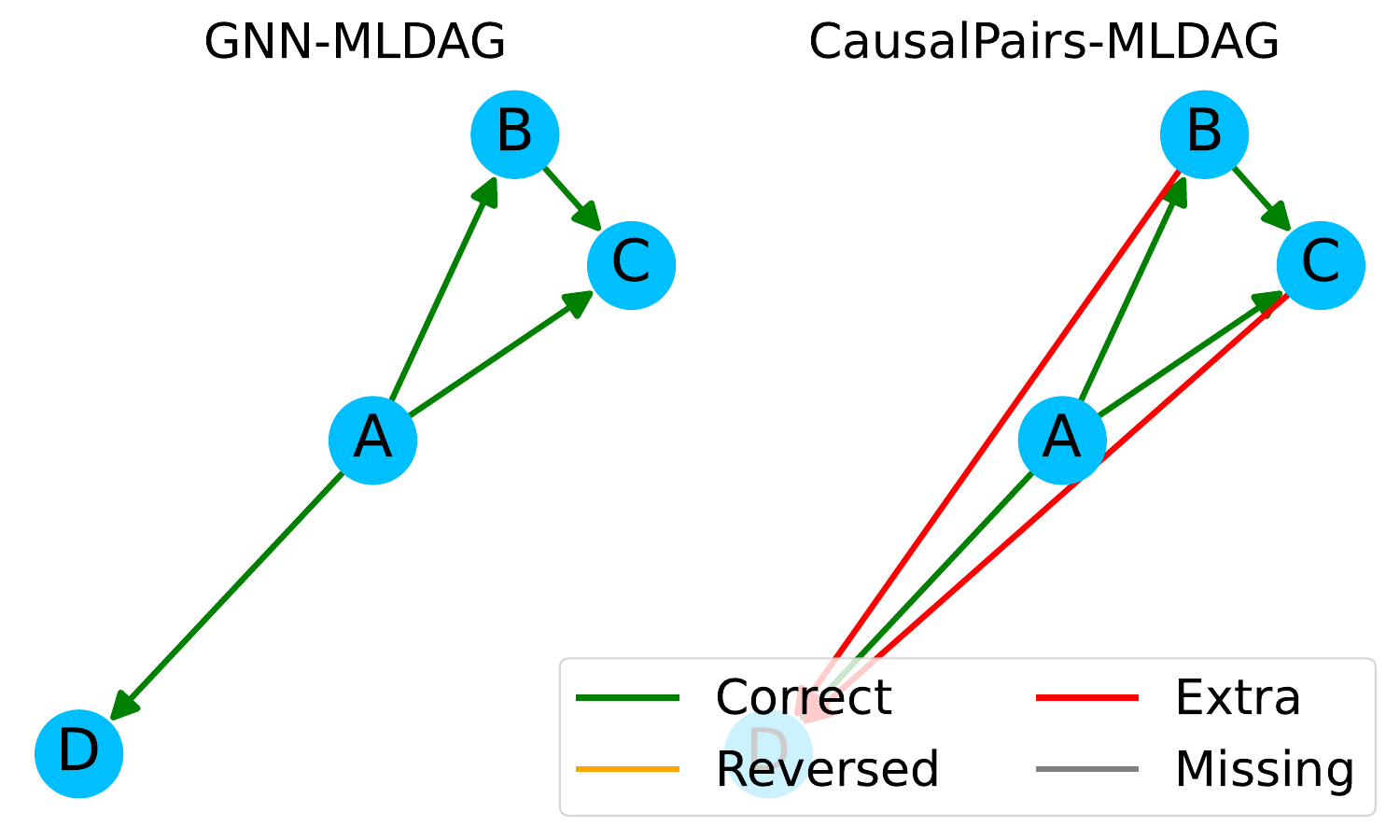}} &
    {\includegraphics[width=.32\linewidth]{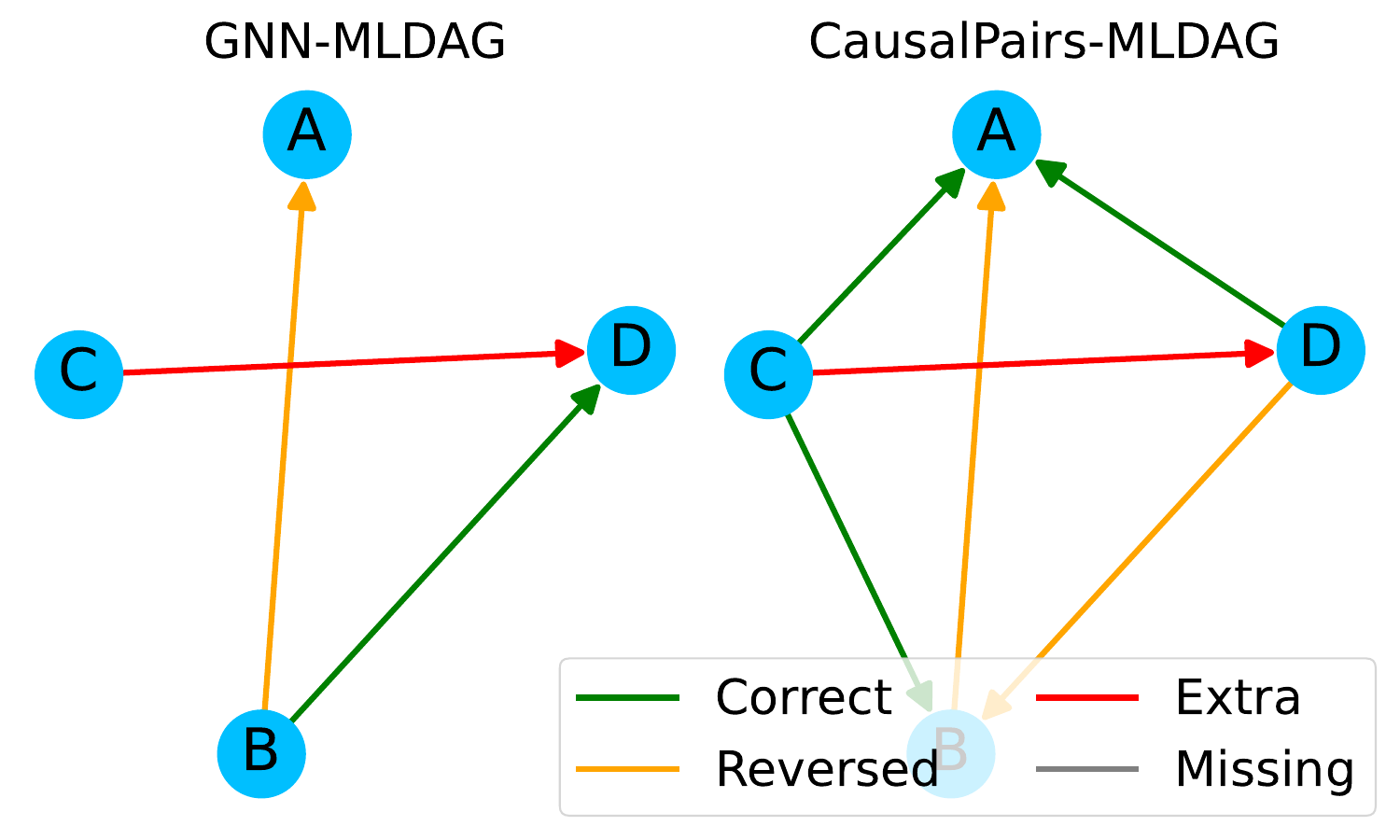}} \\
    
    \small (a) nonlin\_simpson &
    \small (b) symprod\_simpson &
    \small (c) mixed\_simpson
  \end{tabular}

  \caption{Performance comparison between GNN-based methods and CausalPairs methods on smaller CSuite datasets: (a) nonlin\_simpson, (b) symprod\_simpson, and (c) mixed\_simpson. The plots illustrate the number of correct, reversed, extra, and missing edges for each method with respect to the ground truth graphs.}
  \label{fig:csuite_comparison}
\end{figure*}

Notably, as shown in Figure~\ref{fig:csuite_comparison}, our GNN-based methods not only identified the true graph structure but also avoided predicting extraneous edges. In contrast, while CausalPairs methods were able to identify the true edges, they also predicted all possible edges, leading to higher false positives. This underscores the precision of our GNN-based approach in distinguishing true causal relationships from spurious ones.

\begin{table}[htpb]
\centering
\fontsize{9}{11}\selectfont
\caption{Comparison of GNN-based edge probability model (trained on synthetic train data) on the Microsoft CSuite datasets.}
\begin{adjustbox}{width=1\linewidth}
\begin{tabular}{|l|r|r|r|r|r|r|}
\hline
{Dataset Name $\rightarrow$} & \multicolumn{3}{c|}{{large\_backdoor}} & \multicolumn{3}{c|}{{weak\_arrows}} \\
\cline{1-7}
{Method $\downarrow$ | Metrics $\rightarrow$} & {SHD/d} & {TPR} & {FPR} & {SHD/d} & {TPR} & {FPR} \\
\hline
GNN PG & 0.59 & 0.42 & 0.20 & 0.56 & 0.66 & 0.24 \\ \hline
GNN MLG & 0.68 & 0.32 & 0.17 & 0.82 & 0.51 & 0.09 \\ \hline
GNN PDAG & 0.56 & 0.44 & 0.19 & 0.67 & 0.60 & 0.29 \\ \hline
GNN MLDAG & 0.55 & 0.44 & 0.18 & 0.66 & 0.60 & 0.28 \\ \hline
CausalPairs PG & 2.42 & 0.88 & 0.80 & 2.24 & 0.85 & 0.93 \\ \hline
CausalPairs MLG & 1.77 & 0.88 & 0.55 & 1.89 & 0.82 & 0.68 \\ \hline
CausalPairs PDAG & 2.28 & 0.97 & 0.75 & 2.06 & 0.95 & 0.85 \\ \hline
CausalPairs MLDAG & 2.14 & 0.96 & 0.70 & 1.97 & 0.94 & 0.81 \\ \hline
PC & 1.00 & 0.53 & 0.29 & 0.89 & 0.44 & 0.22 \\ \hline
GES & 1.33 & 0.67 & 0.67 & 0.88 & 0.88 & 0.37 \\ \hline
LiNGAM & 2.22 & 0.20 & 0.91 & 1.67 & 0.22 & 0.56 \\ \hline
DAG-GNN & 0.89 & 0.53 & 0.05 & 0.67 & 0.44 & 0.04 \\ \hline
NOTEARS & 1.00 & 0.47 & 0.19 & 0.89 & 0.44 & 0.19 \\ \hline
DiBS & 3.33 & 0.50 & 0.94 & 3.11 & 0.43 & 0.97 \\ \hline
DAGMA & 1.22 & 0.33 & 0.37 & 1.78 & 0.20 & 0.52 \\
\hline
\end{tabular}
\end{adjustbox}
\label{table:csuite}
\end{table}


\begin{table}[htpb]
\centering
\fontsize{9}{11}\selectfont
\caption{Comparison of GNN-based edge probability model (trained on synthetic train data) on the protein network datasets~\citep{sachs2005causal}. DAG-GNN~\citep{yu2019dag} and NOTEARS-MLP~\citep{notears} results for non-standardized data are reported from the original manuscripts.}
\begin{adjustbox}{width=1\linewidth}
\begin{tabular}{|l|r|r|r|r|r|r|}
\hline
{Dataset Type $\rightarrow$} & \multicolumn{3}{c|}{{Standardized}} & \multicolumn{3}{c|}{{Non-standardized}} \\
\cline{1-7}
{Method $\downarrow$ | Metrics $\rightarrow$} & {Predicted} & {Correct} & {Reversed} & {Predicted} & {Correct} & {Reversed} \\
\hline
GNN PG & 19.68 & 6.60 & 6.98 & 19.40 & 5.86 & 7.79 \\ \hline
GNN MLG & 12.07 & 5.13 & 5.64 & 13.81 & 5.48 & 6.86 \\ \hline
GNN PDAG & 17.09 & 6.96 & 5.81 & 16.74 & 4.14 & 8.62 \\ \hline
GNN MLDAG & 14.12 & 6.96 & 5.81 & 12.54 & 4.71 & 7.77 \\ \hline
CausalPairs PG & 36.14 & 6.70 & 7.77 & 38.01 & 6.21 & 8.26 \\ \hline
CausalPairs MLG & 9.82 & 3.04 & 4.26 & 10.41 & 1.52 & 4.04 \\ \hline
CausalPairs PDAG & 33.16 & 7.42 & 6.62 & 34.81 & 6.47 & 7.49 \\ \hline
CausalPairs MLDAG & 18.48 & 4.91 & 5.41 & 20.60 & 4.71 & 6.32 \\ \hline
GES & 34.00 & 5.50 & 9.50 & 34.00 & 5.50 & 9.50 \\ \hline
LiNGAM & 36.00 & 4.00 & 11.00 & 36.00 & 4.00 & 11.00 \\ \hline
DAG-GNN & 6.00 & 1.00 & 5.00 & 18.00 & 8.00 & 3.00 \\ \hline
NOTEARS & 42.33 & 5.83 & 7.18 & 13.00 & 7.00 & 3.00 \\ \hline
DiBS & 46.00 & 7.00 & 7.00 & 50.00 & 8.00 & 9.00 \\ \hline
DAGMA & 11.00 & 3.00 & 5.00 & 7.00 & 5.50 & 1.50 \\
\hline
\end{tabular}
\end{adjustbox}
\label{table:protein}
\end{table}


In Table~\ref{table:protein}, our methods, particularly GNN-PG and GNN-MLDAG, demonstrate strong performance on the real-world protein network dataset, accurately predicting edge counts. Notably, they outperform LiNGAM, DiBS and GES in terms of correct edge predictions, and even match or surpass the performance of recent methods like NOTEARS-MLP, DAG-GNN, and DAGMA. The incorporation of global structural information through GNNs enables accurate edge prediction, while our approach also shows improved directional accuracy, as evident from the lower number of reversed edges achieved by GNN-MLDAG and GNN-PG.

A notable aspect is that DAG-GNN and NOTEARS-MLP exhibit sensitivity to data scaling, with performance variations between standardized and non-standardized data. This sensitivity arises because their continuous optimization processes can be disrupted by changes in data magnitude and distribution. Additionally, LiNGAM, which is designed for non-Gaussian linear models, may struggle with the non-linear relationships present in the protein network dataset. In contrast, our GNN-based methods show consistent performance across both standardized and non-standardized datasets, demonstrating robustness to data scaling. This robustness is attributed to the effective capture and utilization of both local and global structural information by GNNs.

\section{\uppercase{Conclusions}}
\label{conclusions}

In this work, we introduce a probabilistic causal discovery framework that leverages Graph Neural Networks (GNNs) within a supervised learning paradigm. Our approach, trained exclusively on synthetic datasets, effectively generalizes to real-world datasets without requiring additional training.

By exploiting global structural information, our method addresses key limitations of traditional causal discovery techniques, significantly enhancing precision in learning causal graphs. Through integrated node and edge features, our GNN-based model captures complex dependency structures, facilitating more accurate and reliable causal inference. 

Future research directions will include explicitly incorporating acyclicity constraints into the GNN framework to potentially enhance the robustness and accuracy of inferred causal structures. Additionally, investigating advanced GNN architectures may further optimize our method's performance.

\section*{\uppercase{Acknowledgements}}

The research was sponsored by the Army Research Office and was accomplished under Grant Number W911NF-22-1-0035. The views and conclusions contained in this document are those of the authors and should not be interpreted as representing the official policies, either expressed or implied, of the Army Research Office or the U.S. Government. The U.S. Government is authorized to reproduce and distribute reprints for Government purposes notwithstanding any copyright notation herein.

\bibliographystyle{apalike}
{\small
\bibliography{references}
}

\section*{\uppercase{Appendix}}
\label{appendix:features}

\section*{List of Node and Edge Features}

\subsection*{Node Features}
The following features are extracted for each node in the graph, capturing individual statistical properties that are independent of relationships with other nodes.

\begin{itemize}
    \item Min, Max
    \item Numerical Type
    \item Number of Unique Samples
    \item Ratio of Unique Samples
    \item Log of Number of Samples
    \item Normalized Entropy
    \item Normalized Entropy Baseline
    \item Uniform Divergence
    \item Discrete Entropy
    \item Normalized Discrete Entropy
    \item Skewness, Kurtosis
\end{itemize}

\subsection*{Edge Features}

This section provides a comprehensive list of edge features used in our framework, grouped by type, which capture statistical and information-theoretic relationships between pairs of nodes, emphasizing causal relationships or dependencies.

\subsubsection*{Information-Theoretic Features}
\begin{itemize}
    \item \textbf{Mutual Information and Related Measures:}
    \begin{itemize}
        \item Discrete Joint Entropy between nodes
        \item Normalized Discrete Joint Entropy between nodes
        \item Discrete Mutual Information between nodes
        \item Adjusted Mutual Information between nodes
        \item Normalized Discrete Mutual Information
    \end{itemize}
    \item \textbf{Conditional Entropy:}
    \begin{itemize}
        \item Discrete Conditional Entropy for each node pair
    \end{itemize}
    \item \textbf{Divergence Measures:}
    \begin{itemize}
        \item Uniform Divergence for individual nodes
        \item Subtracted Divergence between nodes
    \end{itemize}
\end{itemize}

\subsubsection*{Regression-Based Features}
\begin{itemize}
    \item \textbf{Polynomial Fitting:}
    \begin{itemize}
        \item Polynomial Fit between nodes
        \item Polynomial Fit Error between nodes
        \item Subtracted Polynomial Fit between nodes
    \end{itemize}
    \item \textbf{Error Metrics:}
    \begin{itemize}
        \item Normalized Error Probability for each node pair
        \item Subtracted Normalized Error Probability between nodes
    \end{itemize}
\end{itemize}

\subsubsection*{Statistical Distribution Metrics}
\begin{itemize}
    \item \textbf{Moment-Based Metrics:}
    \begin{itemize}
        \item Second-order moments (Moment21) between nodes
        \item Third-order moments (Moment31) between nodes
        \item Subtracted moments and their absolute values
    \end{itemize}
    \item \textbf{Conditional Distribution Metrics:}
    \begin{itemize}
        \item Entropy variance across node pairs
        \item Skewness variance across node pairs
        \item Kurtosis variance across node pairs
    \end{itemize}
\end{itemize}

\subsubsection*{Correlation Measures}
\begin{itemize}
    \item \textbf{Pearson Correlation:}
    \begin{itemize}
        \item Pearson Correlation Coefficient between nodes
        \item Absolute Pearson Correlation
    \end{itemize}
\end{itemize}

\subsubsection*{Node Pair Comparisons}
\begin{itemize}
    \item \textbf{Sample-Based Comparisons:}
    \begin{itemize}
        \item Maximum, minimum, and difference in the number of unique samples between nodes
    \end{itemize}
    \item \textbf{Entropy Comparisons:}
    \begin{itemize}
        \item Maximum, minimum, and difference in normalized entropy between nodes
    \end{itemize}
\end{itemize}

\subsubsection*{Other Features}
\begin{itemize}
    \item Hilbert-Schmidt Independence Criterion (HSIC) between nodes
    \item Subtracted Information-Geometric Causal Inference (IGCI) values
    \item Absolute differences in kurtosis between nodes
    \item Other advanced metrics derived from normalized probabilities and variance measures
\end{itemize}

\end{document}